\documentclass[11pt]{article}

\usepackage[preprint]{acl}

\usepackage{times}
\usepackage{latexsym}

\usepackage[T1]{fontenc}

\usepackage[utf8]{inputenc}

\usepackage{microtype}

\usepackage{inconsolata}
\usepackage{amsmath}

\usepackage{graphicx}

\usepackage{booktabs} 
\usepackage{multirow} 
\usepackage{makecell} 
\usepackage{array}    
\usepackage{graphicx}    
\usepackage{subcaption}  
\usepackage{enumitem}
\usepackage{colortbl}   
\usepackage{xcolor}     
\usepackage{tcolorbox}          
\usepackage{xcolor}             
\usepackage{pifont}             
\usepackage{graphicx}           
\usepackage{tikz}               

\definecolor{myblue}{RGB}{0, 82, 155}      
\definecolor{mybgblue}{RGB}{200, 220, 255}

\usepackage{xcolor}

\title{Read As Human: Compressing Context via Parallelizable Close Reading and Skimming}

\author{
 \textbf{Jiwei Tang\textsuperscript{1,2}},
 \textbf{Shilei Liu\textsuperscript{2}},
 \textbf{Zhicheng Zhang\textsuperscript{1}},
 \textbf{Qingsong Lv\textsuperscript{1}},
\\
 \textbf{Runsong Zhao\textsuperscript{3}},
 \textbf{Tingwei Lu\textsuperscript{1}},
 \textbf{Langming Liu\textsuperscript{2}},
 \textbf{Haibin Chen\textsuperscript{2}},
\\
 \textbf{Yujin Yuan\textsuperscript{2}},
 \textbf{Hai-Tao Zheng\textsuperscript{1}\thanks{Corresponding authors.}},
 \textbf{Wenbo Su\textsuperscript{2}},
 \textbf{Bo Zheng\textsuperscript{2*}},
\\
 \textsuperscript{1}Tsinghua University \hspace{0.6mm} \textsuperscript{2}Future Living Lab of Alibaba \hspace{0.6mm} \textsuperscript{3}Northeastern University, China
\\
 \texttt{tangjw24@mails.tsinghua.edu.cn} \\
\texttt{zheng.haitao@sz.tsinghua.edu.cn} \\
\texttt{bozheng@alibaba-inc.com}
 \\
}

\begin{document}
\maketitle


\begin{abstract}
Large Language Models (LLMs) demonstrate exceptional capability across diverse tasks. However, their deployment in long-context scenarios is hindered by two challenges: computational inefficiency and redundant information. We propose \textbf{RAM} (\textbf{R}ead \textbf{A}s Hu\textbf{M}an), a context compression framework that adopts an adaptive hybrid reading strategy, to address these challenges. Inspired by human reading behavior (i.e., \textit{close reading} important content while \textit{skimming} less relevant content), \textsc{RAM} partitions the context into segments and encodes them with the input query \textit{in parallel}. High-relevance segments are fully retained (\emph{close reading}), while low-relevance ones are query-guided compressed into compact summary vectors (\emph{skimming}). Both explicit textual segments and implicit summary vectors are concatenated and fed into decoder to achieve both superior performance and natural language format interpretability. To refine the decision boundary between close reading and skimming, we further introduce a contrastive learning objective based on positive and negative query–segment pairs. Experiments demonstrate that \textsc{RAM} outperforms existing baselines on multiple question answering and summarization benchmarks across two backbones, while delivering up to a $12\times$ end-to-end speedup on long inputs (average length 16K; maximum length 32K).
\end{abstract}

\section{Introduction}

\begin{figure}[htb]
\centering
\begin{subfigure}[b]{0.5\textwidth}
    \includegraphics[width=\textwidth]{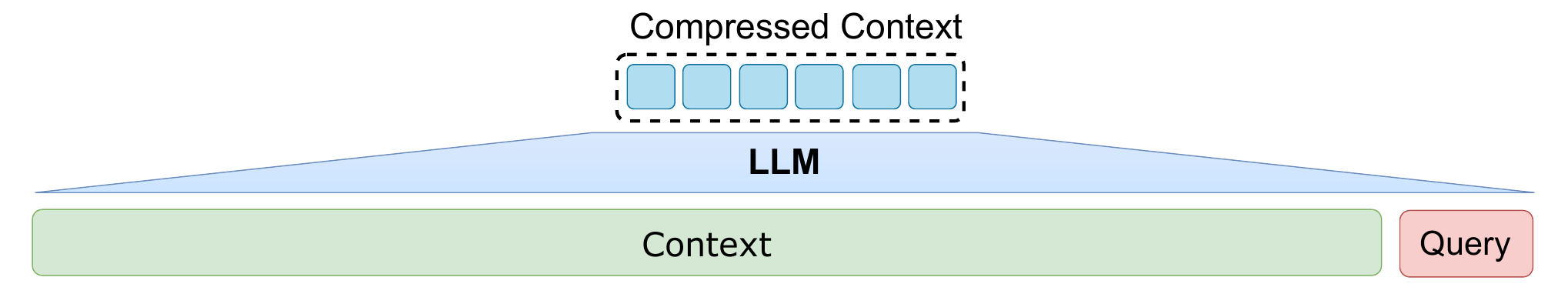}
    \caption{Load all context at once.}
    \label{fig:a}
\end{subfigure}
\hfill
\begin{subfigure}[b]{0.5\textwidth}
    \includegraphics[width=\textwidth]{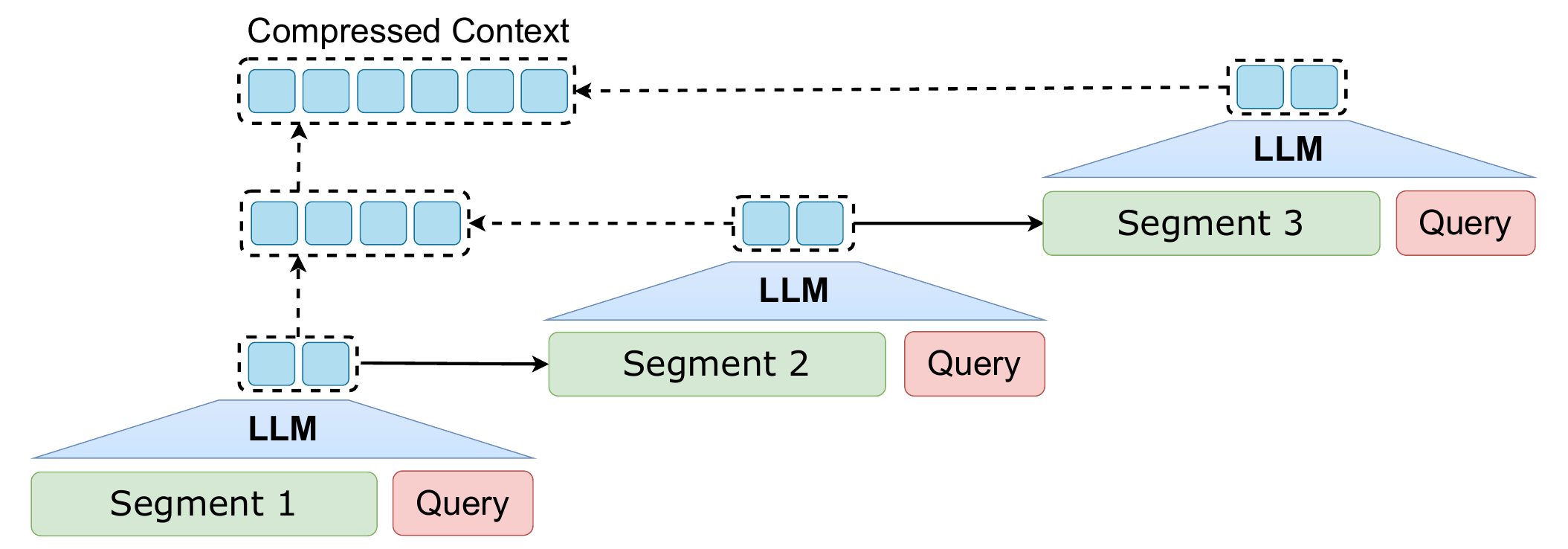}
    \caption{Autoregressive compression.}
    \label{fig:b}
\end{subfigure}
\hfill
\begin{subfigure}[b]{0.5\textwidth}
    \includegraphics[width=\textwidth]{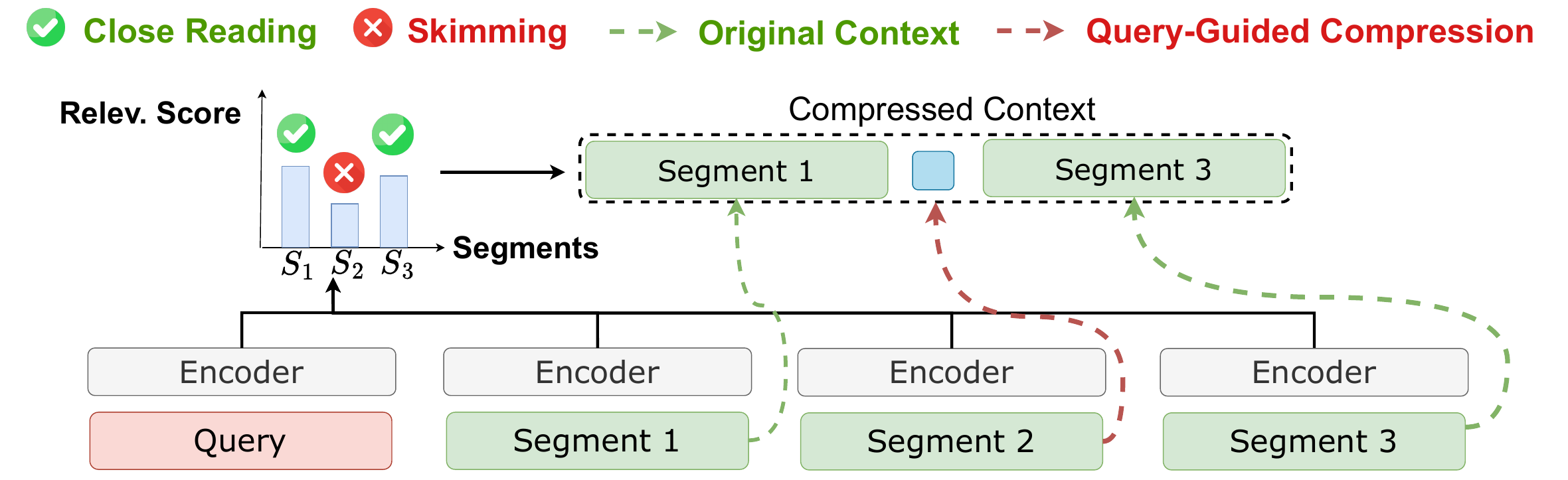}
    \caption{RAM.}
    \label{fig:c}
\end{subfigure}
\caption{Comparison of RAM with other task-aware context compression methods. Existing task-aware methods either require loading the entire input sequence at once for compression (Figure~\ref{fig:a}) or rely on autoregressive compression (Figure~\ref{fig:b}), both of which suffer from computational inefficiency. In contrast, RAM processes all segments and the query \textit{in parallel} and adaptively decides (based on relevance) which segments to \textit{close reading }and which to \textit{skim} (Figure~\ref{fig:c}).}
\label{fig:intro}
\end{figure}

The rise of Large Language Models (LLMs) has profoundly reshaped the paradigm of Natural Language Processing (NLP), demonstrating remarkable generalization across a wide range of tasks~\citep{DBLP:journals/corr/abs-2407-10671,DBLP:journals/corr/abs-2501-12948,kimiteam2025kimik2openagentic,DBLP:journals/corr/abs-2504-07282,DBLP:journals/corr/abs-2511-12913,DBLP:conf/kdd/LiuLY0YZWLWS00025}. However, with the widespread adoption of prompt engineering techniques such as Retrieval-Augmented Generation (RAG)~\citep{DBLP:conf/nips/LewisPPPKGKLYR020}, input prompts are increasingly long, even exceeding tens of thousands of tokens. Deploying LLMs in such long-context scenarios faces two challenges: (1) Computational inefficiency. Mainstream models (e.g., Qwen, DeepSeek) rely on the Transformer architecture~\citep{DBLP:conf/nips/VaswaniSPUJGKP17}, whose self-attention mechanism incurs quadratic time complexity with respect to input sequence length. (2) Information redundancy. Natural language is inherently redundant~\citep{DBLP:journals/bstj/Shannon48}, and this redundancy is exacerbated in long context, thereby degrading performance~\citep{liu2024forgettingcurvereliablemethod,DBLP:conf/acl/JiangWL0L0Q24,DBLP:conf/iclr/00010WWCW24,DBLP:conf/naacl/TangXLZZHZ25,DBLP:journals/corr/abs-2505-12215}.

Context compression has emerged as a promising direction to address these challenges by substantially reducing sequence length while filtering out irrelevant content. Existing methods can fall into two categories: task-agnostic context compression methods~\citep{DBLP:journals/corr/abs-2505-15774,DBLP:conf/acl/LiSC25, DBLP:conf/acl/DaiLHZZWXL25,DBLP:journals/corr/abs-2508-15253,DBLP:conf/iclr/Zhang0XSYD25,DBLP:journals/corr/abs-2505-12215} and task-aware context compression methods. Task-agnostic methods lack query guidance and are prone to losing key information (relevant to query). In contrast, task-aware methods leverage the query to guide compression and better preserve key information. However, existing task-aware methods face two key challenges: (1) Low computational efficiency. As shown in Figure~\ref{fig:intro}, existing methods either process the full long context in at once~\cite{DBLP:conf/acl/CaoCLPHCS24,DBLP:conf/acl/HwangCJSHP25,DBLP:conf/aaai/ZhaoWX25,DBLP:conf/aaai/LiskavetsURKEL25,DBLP:journals/corr/abs-2503-10720}, incurring highly inefficient computation on long sequences, or rely on autoregressive compression~\cite{DBLP:conf/emnlp/YoonLHJK24,DBLP:conf/acl/JiangWL0L0Q24,DBLP:conf/naacl/TangXLZZHZ25}, which also limits computational efficiency. (2) A trade-off between key information retention and interpretability. Some methods \textit{directly prune} low relevance segments or tokens~\cite{DBLP:conf/emnlp/YoonLHJK24,DBLP:conf/acl/JiangWL0L0Q24,DBLP:conf/acl/HwangCJSHP25,DBLP:conf/naacl/TangXLZZHZ25,DBLP:conf/aaai/ZhaoWX25,DBLP:conf/aaai/LiskavetsURKEL25,DBLP:journals/corr/abs-2503-10720}, risking loss of key information, while others compress the context into implicit semantic vectors~\cite{DBLP:conf/acl/CaoCLPHCS24}, sacrificing the interpretability of the natural language format. This motivates the following research question: \emph{How can we achieve efficient context compression that preserves as much key information as possible while remaining natural language format interpretability?}


To address these limitations, we draw inspiration from cognitive science and model human reading behavior: when reading, humans typically perform \emph{close reading} on content highly relevant to their current goal, while adopting a \emph{skimming} strategy for background information~\citep{DBLP:conf/chi/DugganP11,wolf2018reader}. Close reading focuses on the full structural and semantic details of the original context, thereby supporting deep comprehension. In contrast, skimming extracts key information and aggressively discards less relevant content, significantly reducing cognitive load while still capturing the query-relevant semantic gist. Motivated by this, we propose \textbf{RAM} (\textbf{R}ead \textbf{A}s Hu\textbf{M}an), which formalizes context compression as an efficient and adaptive hybrid reading strategy. Specifically, \textsc{RAM} first partitions a long context into multiple segments and processes all segments together with the input query \textit{in parallel}, \textit{avoiding the efficiency bottlenecks of existing methods that either load the entire context at once or rely on autoregressive compression}. Subsequently, RAM adaptively compresses the context based on query-segment relevance: highly relevant segments are fully retained (i.e., \emph{close reading}) to ensure key information is preserved in natural language format, while less relevant segments are compressed via a query-guided mechanism into compact, implicit summary vectors (i.e., \emph{skimming}), capturing only the query-relevant semantic gist while drastically reducing less relevant content. Then, RAM concatenates the explicit \emph{close reading} segments with the implicit \emph{skimming} summary vectors to form a hybrid contextual representation, which is fed into the decoder. \emph{This design not only preserve as much key information as possible but also maintains natural language format interpretability.} We further introduce a contrastive learning objective that leverages annotated positive and negative query-segments pairs, optimizing the RAM’s ability to distinguish between \emph{close reading} and \emph{skimming}, which enables \textsc{RAM} to more faithfully emulate human-like adaptive reading.

Our main contributions are threefold: (1) We propose \textsc{RAM}, an efficient and interpretable compression framework that combines explicit and implicit representations. By parallelizing segment encoding with the query and applying differentiated treatment based on relevance, \textsc{RAM} avoids the inefficiencies of full-sequence loading or autoregressive compression while maintaining interpretability. (2) We introduce a contrastive learning object for more adaptive compression. By modeling query-segment relevance as a contrastive task, the model learns a more robust decision boundary between \textit{close reading} and \textit{skimming}. (3) RAM achieves superior performance across multiple QA and summarization benchmarks under various compression budgets across two backbones, offers approximately 12$\times$ end-to-end speedup on long inputs (average length 16K; maximum length 32K).

\begin{figure*}[htb]
    \centering
    \includegraphics[width=1\linewidth]{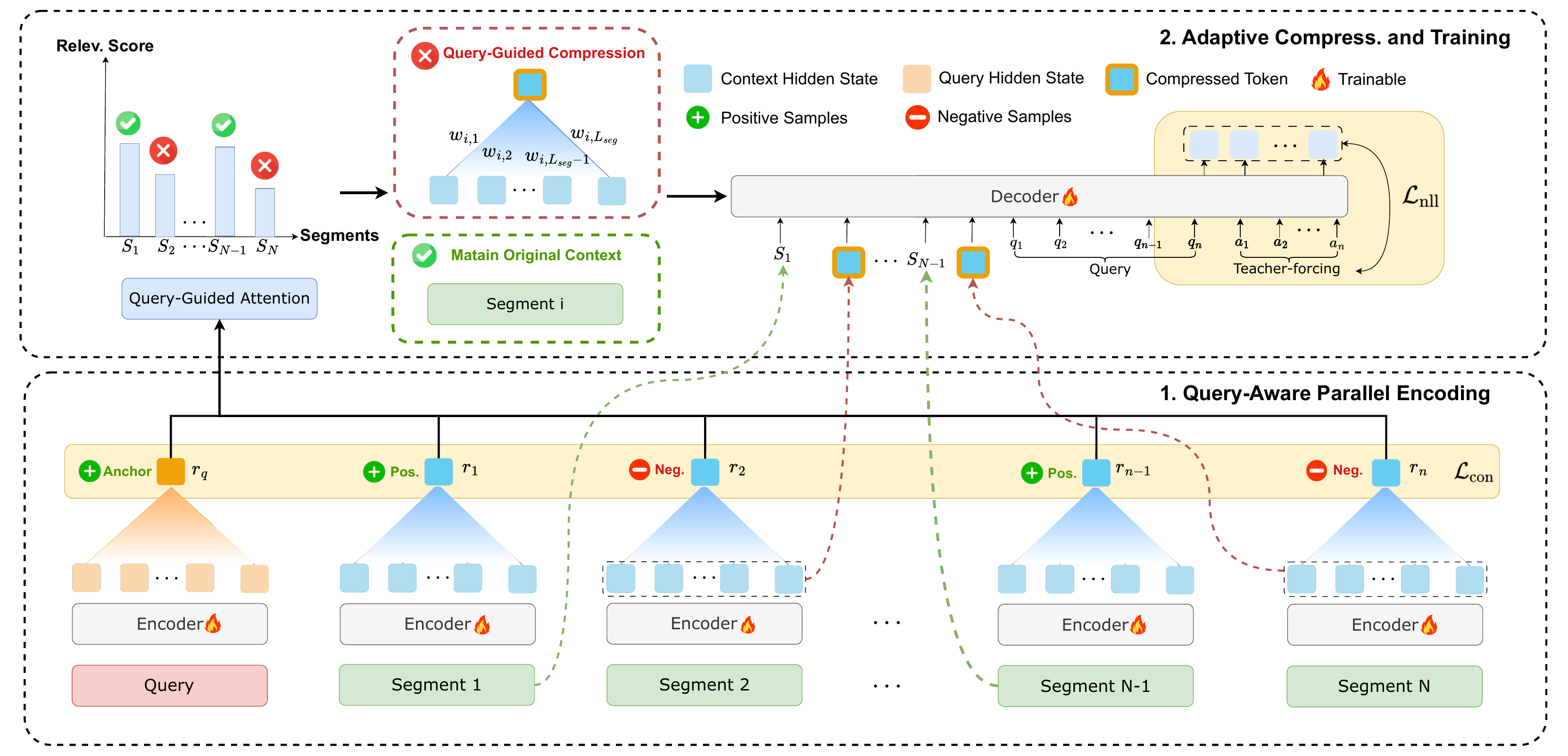}
    \caption{Overview of the RAM framework. The framework consists of two main stages: (1) \textbf{Query-Aware Parallel Encoding}: The query and segmented context are encoded \textit{in parallel}. A query-guided attention mechanism computes a relevance score for each segment, determining whether to retain it as original text (\textit{close reading}) or compress it into a compact vector (\textit{skimming}), with the number of segments to retain derived from the compression ratio via Eq.~\eqref{eq:k_def}. (2) \textbf{Adaptive Compression and Training}: The retained text and compressed vectors are fed into a decoder to produce the final output. The model is trained end-to-end using a language modeling objective $\mathcal{L}_{\text{nll}}$ and a contrastive learning objective $\mathcal{L}_{\text{con}}$ to jointly optimize overall performance.}
    \label{fig:framework}
    \vspace{-1em}
\end{figure*}

\section{Related Work}
\paragraph{Task-Agnostic Context Compression Methods.}
Task-agnostic methods compress input context without relying on specific queries, preserving global semantics to support diverse downstream tasks. This category of work can further fall into two categories: (1) Hard prompt compression, which retains explicit textual tokens. Representative methods include metric-based pruning using information entropy~\citep{DBLP:conf/emnlp/0001DGL23, DBLP:conf/emnlp/JiangWLYQ23} or bidirectional semantics~\citep{DBLP:conf/acl/PanWJXLZLR0LZQ024}, and summarization-based approaches~\citep{DBLP:conf/iclr/XuSC24} that train a query-agnostic summarizer. (2) Soft prompt compression, which maps context into non-lexical soft embeddings. Representatives include encoder-decoder frameworks~\citep{DBLP:conf/iclr/00010WWCW24, DBLP:conf/nips/0002W00CWZ024,DBLP:journals/corr/abs-2407-09252,DBLP:conf/emnlp/TanLPWZK0P24, DBLP:journals/corr/abs-2505-15774,DBLP:conf/acl/LiSC25, DBLP:conf/acl/DaiLHZZWXL25,DBLP:journals/corr/abs-2508-15253,DBLP:journals/corr/abs-2505-12215,zhao2025positionidsmatterenhanced,liu2025autoencodingfreecontextcompressionllms}, attention mask modification~\citep{DBLP:conf/nips/Mu0G23, DBLP:journals/corr/abs-2504-08934,DBLP:conf/cvpr/YeGHGT25,li2025admtreecompressinglengthycontext}, and autoregressive modeling~\citep{DBLP:conf/emnlp/ChevalierWAC23, DBLP:conf/iclr/Zhang0XSYD25,deng2025unigist,chen2025dastcontextawarecompressionllms} that treats compression as sequence generation conditioned on previously compressed tokens. \textit{Despite preserving general semantics, these methods prone to discard task-relevant information, degrading performance due to the lack of query awareness.}
\vspace{-1em}
\paragraph{Task-Aware Context Compression Methods.}
Task-aware methods incorporate the query during compression to retain task-relevant content. Representative methods include generating query-guided summaries~\citep{DBLP:conf/emnlp/YoonLHJK24,DBLP:conf/acl/HwangCJSHP25}, merging query-weighted tokens~\citep{DBLP:conf/acl/CaoCLPHCS24}, or pruning low query relevance tokens~\citep{DBLP:conf/acl/JiangWL0L0Q24, DBLP:conf/naacl/TangXLZZHZ25,DBLP:conf/aaai/LiskavetsURKEL25,DBLP:journals/corr/abs-2503-10720,DBLP:conf/aaai/ZhaoWX25}. \textit{While preserving task-specific content, these methods either require loading the full sequence at once or depend on iterative autoregressive compression (both significantly hindering efficiency) and face an inherent trade-off between preserving key information and maintaining natural language format interpretability.}

\section{Method}

We propose \textsc{RAM}, a novel context compression framework that mimics human reading behavior: close reading highly relevant segments while skimming over less relevant ones via query-guided compression. The method operates within an encoder–decoder architecture and consists of two core stages: (1) query-aware parallel encoding of context segments, and (2) adaptive compression and training. Below, we detail each component.

\subsection{Query-Aware Parallel Encoding}

Given a query $Q$ and a long context $\mathbf{C}$ segmented into $N$ \emph{equal-length} chunks $\{S_1, S_2, \dots, S_N\}$., \textsc{RAM} processes all segments \textit{in parallel} with the query using a shared encoder, avoiding the quadratic cost of full-sequence attention and iterative compression.

\paragraph{Parallel Encoding.}
The original input sequence $\{S_1, S_2, \dots, S_N, Q\}$ is passed through a trainable LLM-based encoder (e.g., LLaMA or Qwen) \textit{in parallel} to obtain contextualized hidden states. Let $\mathbf{H}$ denote the last hidden states.

\paragraph{Representative Tokens for the Query and Segments.}
We construct a compact representation for each text sequence by mean-pooling its last hidden states. For segment $S_i$ with length $L_{\text{seg}}$, its representation is computed as
\begin{equation}
\mathbf{r}_i \;=\; \frac{1}{L_{\text{seg}}}\sum_{j=1}^{L_{\text{seg}}}\mathbf{h}_{i,j},
\end{equation}
where $L_{seg}$ refers to the token length of each segment; $\mathbf{h}_{i,j}$ denotes the last hidden state of the $j$-th token in $S_i$. 
The query representation $\mathbf{r}_q$ is obtained in the same way by mean-pooling the last-layer hidden states of the query tokens.

\subsection{Adaptive Compression and Training}

Based on relevance to the query, \textsc{RAM} dynamically decides whether to preserve each segment verbatim (``close reading'') or compress it into a single vector (``skimming''), guided by a learnable selection mechanism.

\paragraph{Query-Guided Attention.}
We compute cosine similarity between $\mathbf{r}_q$ and each $\mathbf{r}_i$, followed by a softmax to obtain segment probabilities $p_i$: 
\begin{equation}
    p_i = \frac{\exp\left(\mathbf{r}_q^\top \mathbf{r}_i / \tau\right)}{\sum_{j} \exp\left(\mathbf{r}_q^\top \mathbf{r}_j / \tau\right)} \, .
\end{equation}
Given a sampled compression rate $\alpha$ (i.e., sampled from $\{2, 4, 8, 16, 32\}$), we determine the number of segments $k$ to fully preserve. $k$ satisfies:
\begin{equation}
    k = \lfloor \frac{L_{\mathrm{org}}}{\alpha L_{\mathrm{seg}}} \rfloor \, ,
    \label{eq:k_def}
\end{equation}
where $L_{org}$ is the token length of original input. We select the top-$k$ segments with highest $p_i$ for retention; others are compressed.

\paragraph{Query-Guided Compression (Skimming).}
For segments marked for skimming, we compute a query-aware weighted average of token hidden states:
\begin{equation}
\begin{aligned}
w_{i,t} &= \mathrm{softmax}_{t}\big(\operatorname{cos}(\mathbf{h}_{i,t}, \mathbf{r}_q)\big), \\
\mathbf{c}_i &= \sum_{t=1}^{L_{\text{seg}}} w_{i,t}\,\mathbf{h}_{i,t}.
\end{aligned}
\end{equation}
where $\operatorname{cos}(\cdot,\cdot)$ denotes cosine similarity.
This yields a single embedding $\mathbf{c}_i \in \mathbf{R}^d$ per skimmed segment, where $d$ is the hidden dimension.

\paragraph{Hybrid Compressed Representation Construction.}
Let $\mathcal{K} \subseteq \{1, \dots, N\}$ denote the set of segment indices selected for close reading, and let $\bar{\mathcal{K}}$ be its complement. For each segment $i$, we construct its contribution to the compressed context representation as
\begin{equation}
\tilde{\mathbf{M}}_i = \mathbf{I}_{\{i \in \mathcal{K}\}} \cdot \mathbf{e}(S_i) + \mathbf{I}_{\{i \in \bar{\mathcal{K}}\}} \cdot \mathbf{W}_{\text{align}} \mathbf{c}_i,
\end{equation}
where $\mathbf{e}(\cdot)$ denotes the word embedding lookup in
the Decoder LLM; $\mathbf{c}_i \in \mathbf{R}^{d}$ is the query-guided compressed vector obtained via query-guided compression (skimming), and $\mathbf{W}_{\text{align}} \in \mathbf{R}^{d \times d}$ is a trainable semantic alignment matrix applied only to skimmed segments to bridge the representation gap between explicit and implicit forms. The indicator $\mathbf{I}_{\{\cdot\}}$ selects the appropriate representation path per segment. The final compressed context is then formed by concatenating all segment representations in their original order:
\begin{equation}
\mathbf{M} = \big[ \tilde{\mathbf{M}}_1; \tilde{\mathbf{M}}_2; \dots; \tilde{\mathbf{M}}_N \big] \in \mathbf{R}^{L_{\text{c}} \times d},
\end{equation}
where $L_{\text{c}} = \sum_{i \in \mathcal{K}} L_{\text{seg}} + |\bar{\mathcal{K}}|$ is the total length after compression. This hybrid representation preserves verbatim tokens for high-relevance segments while replacing low-relevance ones with compact, query-guided skimming, thereby achieving both fidelity and efficiency.

\paragraph{Training Objective.}
The compressed memory $\mathbf{M}$ is concatenated with the query embeddings and answer tokens, and fed into the decoder for standard teacher-forced language modeling. Let $\mathbf{Y} = [y_1, \dots, y_{N_a}]$ denotes the answer token sequence. The language modeling loss is
\begin{equation}
\mathcal{L}_{\text{nll}} = -\frac{1}{N_a} \sum_{t=1}^{N_a} \log P(y_t \mid y_{<t}, \mathbf{M}, Q),
\end{equation}
where $P(y_t \mid \cdot)$ is derived from the softmax-normalized logits, and loss is computed only over answer positions.

To improve relevance discrimination, we introduce a contrastive loss during training. Given ground-truth positive/negative segment labels (e.g., from answer span annotations), we encourage higher similarity between the query and positive segments, and lower similarity with negative ones:
\begin{equation}
\mathcal{L}_{\text{con}} = -\frac{1}{|\mathcal{P}|} \sum_{i \in \mathcal{P}} \log \frac{\exp(\mathrm{cos}(\mathbf{r}_q, \mathbf{r}_i)/\tau)}{\sum_{j=1}^N \exp(\mathrm{cos}(\mathbf{r}_q, \mathbf{r}_j)/\tau)},
\end{equation}
where $\mathcal{P}$ is the set of positive segment indices and $\tau$ is a temperature parameter.

The total training objective combines both terms:
\begin{equation}
\mathcal{L}_{\text{RAM}} = \mathcal{L}_{\text{nll}} + \mathcal{L}_{\text{con}}.
\end{equation}
This design enables \textsc{RAM} to achieve high compression efficiency, strong task performance, and natural language format interpretability.

\begin{table*}[htb]
\centering
\fontsize{8}{9}\selectfont
\caption{Experimental results on four QA benchmarks and the MultiNews summarization benchmark. Closed-book indicates using only the input question as the input, while Original Prompt indicates using original context as the input. We \textbf{bold} the optimal results.}
\label{tab:overall_comparison}
\begin{tabular}{l | c c c c c c c c c | c c}
\toprule
\multirow{2}{*}{\textbf{Methods}} & 
  \multicolumn{2}{c}{\textbf{NaturalQA}} & 
  \multicolumn{2}{c}{\textbf{2WikiMQA}} & 
  \multicolumn{2}{c}{\textbf{HotpotQA}} & 
  \multicolumn{2}{c}{\textbf{NarrativeQA}} &
  \textbf{MultiNews} &
  \multicolumn{2}{c}{\textbf{AVG}} \\
\cmidrule(lr){2-3} \cmidrule(lr){4-5} \cmidrule(lr){6-7} \cmidrule(lr){8-9} \cmidrule(lr){10-10} \cmidrule(lr){11-12}
& \textbf{EM} & \textbf{F1} & \textbf{EM} & \textbf{F1} & \textbf{EM} & \textbf{F1} & \textbf{EM} & \textbf{F1} & \textbf{F1} & \textbf{EM} & \textbf{F1} \\
\midrule
\multicolumn{12}{c}{\textbf{LLaMA-3.1-8B-Instruct}} \\
\midrule
Closed-book & 14.05 & 21.98 & 17.14 & 23.42 & 10.77 & 19.21 & 1.03 & 9.00 & - & 10.75 & 18.40 \\
Original Prompt & 37.63 & 48.25 & 26.07 & 39.60 & 26.81 & 45.86 & 6.39 & 19.26 & 34.79 & 24.22 & 37.55 \\
\midrule
\multicolumn{12}{c}{\textit{4x Compression Constraint}} \\
\midrule
ICAE & 35.82 & 37.57 & 32.66 & 37.78 & 26.18 & 36.17 & 4.70 & 12.46 & 28.16 & 24.84 & 30.43 \\
LLMLingua-2-large & 29.98 & 43.92 & 18.16 & 32.76 & 29.83 & 45.25 & 12.59 & 24.93 & \textbf{30.08} & 22.64 & 35.39 \\
Activation Beacon & 47.04 & 58.97 & 36.69 & 44.20 & 43.78 & 57.44 & 18.61 & 28.29 & 28.52 & 36.53 & 43.48 \\
Provence & 37.25 & 49.13 & 30.25 & 45.87 & 40.57 & 57.51 & 15.13 & 30.11 & 30.14 & 30.80 & 42.55 \\
EXIT & 43.31 & 57.70 & 28.43 & 42.25 & 45.55 & 58.81 & 11.00 & 24.17 & 29.71 & 32.07 & 42.53 \\
LongLLMLingua & 45.50 & 58.34 & 24.58 & 35.30 & 34.88 & 51.40 & 7.15 & 19.17 & 26.32 & 28.03 & 38.11 \\
\midrule
{\cellcolor[rgb]{0.925,0.957,1}}\textbf{RAM} & {\cellcolor[rgb]{0.925,0.957,1}}\textbf{65.35} & {\cellcolor[rgb]{0.925,0.957,1}}\textbf{59.22} & {\cellcolor[rgb]{0.925,0.957,1}}\textbf{48.66} & {\cellcolor[rgb]{0.925,0.957,1}}\textbf{54.89} & {\cellcolor[rgb]{0.925,0.957,1}}\textbf{46.24} & {\cellcolor[rgb]{0.925,0.957,1}}\textbf{59.13} & {\cellcolor[rgb]{0.925,0.957,1}}\textbf{19.64} & {\cellcolor[rgb]{0.925,0.957,1}}\textbf{32.47} & {\cellcolor[rgb]{0.925,0.957,1}}29.21 & {\cellcolor[rgb]{0.925,0.957,1}}\textbf{44.97} & {\cellcolor[rgb]{0.925,0.957,1}}\textbf{46.98} \\
\midrule
\multicolumn{12}{c}{\textit{8x Compression Constraint}} \\
\midrule
ICAE & 36.13 & 38.31 & 33.42 & 37.78 & 26.92 & 35.77 & 4.41 & 12.41 & \textbf{27.92} & 25.22 & 30.44 \\
LLMLingua-2-large & 20.11 & 33.58 & 14.05 & 27.62 & 21.29 & 34.55 & 11.28 & 22.46 & 27.55 & 16.68 & 29.15 \\
Activation Beacon & 41.21 & 55.09 & 35.53 & 42.98 & 36.29 & 48.57 & 17.95 & 26.80 & 25.19 & 32.74 & 39.73 \\
Provence & 34.16 & 47.19 & 28.81 & 43.12 & 37.22 & 49.91 & 14.10 & 29.15 & 26.19 & 28.57 & 39.11 \\
EXIT & 43.88 & 55.42 & 22.94 & 33.33 & 32.16 & 50.19 & 6.86 & 18.62 & 27.15 & 26.46 & 36.94 \\
LongLLMLingua & 39.50 & 54.30 & 20.33 & 30.13 & 28.63 & 44.10 & 4.52 & 15.84 & 21.56 & 23.24 & 33.19 \\
\midrule
{\cellcolor[rgb]{0.925,0.957,1}}\textbf{RAM} & {\cellcolor[rgb]{0.925,0.957,1}}\textbf{62.41} & {\cellcolor[rgb]{0.925,0.957,1}}\textbf{57.14} & {\cellcolor[rgb]{0.925,0.957,1}}\textbf{39.63} & {\cellcolor[rgb]{0.925,0.957,1}}\textbf{45.24} & {\cellcolor[rgb]{0.925,0.957,1}}\textbf{38.82} & {\cellcolor[rgb]{0.925,0.957,1}}\textbf{50.80} & {\cellcolor[rgb]{0.925,0.957,1}}\textbf{18.80} & {\cellcolor[rgb]{0.925,0.957,1}}\textbf{31.57} & {\cellcolor[rgb]{0.925,0.957,1}}26.11 & {\cellcolor[rgb]{0.925,0.957,1}}\textbf{39.92} & {\cellcolor[rgb]{0.925,0.957,1}}\textbf{42.17} \\
\midrule
\multicolumn{12}{c}{\textbf{Qwen3-4B-Instruct}} \\
\midrule
Closed-book & 10.17 & 17.75 & 13.67 & 23.92 & 12.16 & 20.45 & 0.47 & 9.65 & - & 9.12 & 17.94 \\
Original Prompt & 32.77 & 44.44 & 32.94 & 43.60 & 44.33 & 60.47 & 8.55 & 20.06 & 32.09 & 29.65 & 40.13 \\
\midrule
\multicolumn{12}{c}{\textit{4x Compression Constraint}} \\
\midrule
ICAE & 18.97 & 20.05 & 25.91 & 28.52 & 18.41 & 25.34 & 2.53 & 11.72 & 22.78 & 16.45 & 21.68 \\
LLMLingua-2-large & 23.09 & 35.72 & 25.17 & 31.58 & 28.20 & 41.02 & 7.89 & 19.03 & 29.56 & 21.09 & 31.38 \\
Provence & 31.11 & 43.39 & 39.52 & 48.32 & 42.38 & 56.11 & 11.00 & 24.46 & 28.30 & 31.00 & 40.12 \\
EXIT & 40.00 & 52.86 & 37.06 & 45.04 & 45.24 & 58.13 & 5.64 & 15.59 & 31.86 & 31.99 & 40.70 \\
LongLLMLingua & 40.23 & 53.01 & 26.81 & 32.68 & 30.13 & 42.86 & 2.82 & 12.20 & 24.15 & 25.00 & 32.98 \\
\midrule
{\cellcolor[rgb]{0.925,0.957,1}}\textbf{RAM} & {\cellcolor[rgb]{0.925,0.957,1}}\textbf{66.59} & {\cellcolor[rgb]{0.925,0.957,1}}\textbf{59.97} & {\cellcolor[rgb]{0.925,0.957,1}}\textbf{50.39} & {\cellcolor[rgb]{0.925,0.957,1}}\textbf{56.47} & {\cellcolor[rgb]{0.925,0.957,1}}\textbf{46.37} & {\cellcolor[rgb]{0.925,0.957,1}}\textbf{59.28} & {\cellcolor[rgb]{0.925,0.957,1}}\textbf{19.45} & {\cellcolor[rgb]{0.925,0.957,1}}\textbf{31.92} & {\cellcolor[rgb]{0.925,0.957,1}}\textbf{32.07} & {\cellcolor[rgb]{0.925,0.957,1}}\textbf{45.70} & {\cellcolor[rgb]{0.925,0.957,1}}\textbf{47.94} \\
\midrule
\multicolumn{12}{c}{\textit{8x Compression Constraint}} \\
\midrule
ICAE & 19.49 & 19.71 & 26.07 & 28.55 & 18.08 & 25.84 & 2.45 & 11.75 & 23.42 & 16.52 & 21.85 \\
LLMLingua-2-large & 15.37 & 26.67 & 21.37 & 25.57 & 18.60 & 28.23 & 6.20 & 15.99 & 26.41 & 15.39 & 24.57 \\
Provence & 31.30 & 43.01 & 37.25 & 44.72 & 36.74 & 48.71 & 10.71 & 24.34 & 24.47 & 29.00 & 37.05 \\
EXIT & 41.54 & 53.90 & 29.80 & 35.59 & 35.75 & 48.40 & 2.82 & 11.53 & 28.07 & 27.48 & 35.50 \\
LongLLMLingua & 31.41 & 45.27 & 23.65 & 27.91 & 24.45 & 35.51 & 1.88 & 9.76 & 20.48 & 20.35 & 27.79 \\
\midrule
{\cellcolor[rgb]{0.925,0.957,1}}\textbf{RAM} & {\cellcolor[rgb]{0.925,0.957,1}}\textbf{61.09} & {\cellcolor[rgb]{0.925,0.957,1}}\textbf{56.15} & {\cellcolor[rgb]{0.925,0.957,1}}\textbf{40.24} & {\cellcolor[rgb]{0.925,0.957,1}}\textbf{45.66} & {\cellcolor[rgb]{0.925,0.957,1}}\textbf{37.90} & {\cellcolor[rgb]{0.925,0.957,1}}\textbf{49.15} & {\cellcolor[rgb]{0.925,0.957,1}}\textbf{19.92} & {\cellcolor[rgb]{0.925,0.957,1}}\textbf{31.22} & {\cellcolor[rgb]{0.925,0.957,1}}\textbf{28.50} & {\cellcolor[rgb]{0.925,0.957,1}}\textbf{39.79} & {\cellcolor[rgb]{0.925,0.957,1}}\textbf{42.14} \\
\bottomrule
\end{tabular}
\end{table*}

\section{Experiment}

In this section, we aim to address the following four research questions (RQs): (1) How does RAM perform compared to baseline methods (RQ1)? (2) How efficient is RAM (RQ2)? (3) How efficient and robust is RAM under various compression ratios in long-context scenarios (RQ3)? (4) How effective is each component within RAM (RQ4)?

\subsection{Settings}

\paragraph{Training.} RAM requires \textit{only a single training} to support diverse downstream tasks under various compression ratios, including question answering and summarization. We randomly sample 20,000 examples from each of HotpotQA~\citep{DBLP:conf/emnlp/Yang0ZBCSM18}, 2WikiMQA~\citep{DBLP:conf/coling/HoNSA20}, NaturalQuestions~\citep{DBLP:journals/tacl/LiuLHPBPL24}, NarrativeQA~\citep{DBLP:journals/tacl/KociskySBDHMG18}, and MultiNews~\citep{DBLP:conf/acl/FabbriLSLR19} to construct the final training set (more details about datasets in Appendix~\ref{apx:dataset}). All training samples are truncated to a maximum token length of 20K. During training, we use a batch size of 32 and a learning rate of $1 \times 10^{-5}$ with a linear learning rate decay schedule. We set the segment size to 50 to enable finer-grained control over compression; in practice, the segment size has minimal impact on performance (see Appendix~\ref{apx:impact_of_seg}). Following~\citet{DBLP:conf/icml/ChenK0H20}, $\tau$ is set to 0.1. The training paradigm is illustrated in Figure~\ref{fig:framework}. For each training sample, we randomly sample a compression rate from the $\{2, 4, 8, 16, 32\}$ to cover various compression situations. We set the maximum test length for NarrativeQA to 32K, resulting in a test set with an average length of 16K.

\paragraph{Contrastive Learning Data Processing.}
We convert ground truth positions from NaturalQuestions, HotpotQA, and 2WikiMQA into segments. MultiNews is excluded as it is a summarization dataset. For NarrativeQA (which lacks position annotations), we segment documents and use Qwen3-235B-A22B-Instruct to label segments as answer-containing or not (see Appendix~\ref{apx:prompt_temp}).


\paragraph{Implementation Details.} Our implementation is based on LLaMA-3.1-8B (Instruct) and Qwen3-4B (Instruct). To ensure a fair comparison, all baseline results are reproduced using the officially released code. All experiments are conducted on 8 GPUs with compute performance comparable to NVIDIA H800, using the Hugging Face framework.


\paragraph{Evaluation Metrics.} Following~\citet{hwang-etal-2025-exit}, for question answering tasks, we report Exact Match (EM) and F1 score~\citep{DBLP:conf/emnlp/Yang0ZBCSM18}. For summarization on MultiNews, performance is measured using the F1 score.


\paragraph{Baselines.} We conduct comprehensive comparisons against both task-specific and task-agnostic text compression methods. Specifically, we include task-specific approaches (e.g., LongLLMLingua~\citep{DBLP:conf/acl/JiangWL0L0Q24}, Provence~\citep{DBLP:conf/iclr/ChirkovaFNC25}, EXIT~\citep{hwang-etal-2025-exit}) and task-agnostic approaches (e.g., LLMLingua-2~\citep{DBLP:conf/acl/PanWJXLZLR0LZQ024}, Activation Beacon~\citep{DBLP:conf/iclr/Zhang0XSYD25}). We compare with Activation Beacon under the setting using LLaMA-3.1-8B-Instruct as the backbone, as its open-source code only supports LLaMA-3.1-8B-Instruct. Additionally, we also report model performance under the settings of the original (uncompressed) prompt and zero-shot prompting.

\subsection{Main Results (RQ1)}
As shown in Table~\ref{tab:overall_comparison}, RAM demonstrates significant advantages across multiple benchmarks. We summarize our findings as follows: 
(1) \textbf{Performance Superiority.} Under both 4$\times$ and 8$\times$ compression constraints, RAM consistently outperforms existing baselines on all benchmarks, achieving state-of-the-art results (bolded) on EM and F1 metrics most of the time. This indicates its superior capability in preserving semantics and supporting question answering under long-context compression scenarios.  
(2) \textbf{Robustness.} RAM maintains stable and strong performance across different backbone models (\textit{i.e.}, LLaMA-3.1-8B-Instruct and Qwen3-4B-Instruct) and compression constraint (\textit{i.e.}, 4$\times$ and 8$\times$). This confirms the adaptability of RAM to diverse model architectures and compression strategies.  
(3) \textbf{Length Extrapolation Capability}: Although trained with a maximum input length of 20K tokens, RAM outperforms both the Original Prompt and other baselines on NarrativeQA (with the max length up to 32K tokens), demonstrating not only effective context compression but also the ability to maintain semantic coherence and inference accuracy on substantially longer inputs, highlighting its strong generalization and extrapolation capacity. This extrapolation behavior is particularly promising for real-world applications where input lengths often exceed training regimes, suggesting that RAM learns compositional representations rather than memorizing fixed-length patterns.

\subsection{Efficiency Analysis (RQ2)}
\label{subsec:efficiency}

We analyze the computational efficiency of the RAM framework. By segmenting the long context and encoding each segment with the query \textit{in parallel} (followed by ``skimming'' low-relevance segments via lightweight compression). RAM \emph{fundamentally avoids} the bottlenecks of conventional approaches that either process the full input in one shot or rely on autoregressive iteration. This design drastically reduces inference cost. The entire pipeline consists of two stages: (1) compression, which includes query-aware parallel encoding and adaptive compression, and (2) decoding, whose floating-point operations (FLOPs) can be modeled separately.

\paragraph{Compression Stage.} This stage involves two core components. First, the original context $\mathbf{C}$ of length $L_{\text{org}}$ is split into $N$ segments, each encoded \textit{in parallel} alongside the query $Q$ (length $L_q$). Second, a query-guided attention mechanism computes a relevance score for each segment to decide whether to retain it as-is (``close reading'') or compress it into a compact vector (``skimming''). Compressed segments are aggregated via a lightweight operation (e.g., weighted average) into a single summary vector. Let $L_c$ denote the total effective length fed to the decoder (i.e., sum of retained token lengths and number of skimmed vectors), and $\alpha = L_c / L_{\text{org}}$ the compression ratio. The FLOPs of this stage are:
\begin{table}[htb]
\scriptsize
\centering
\caption{Latency evaluation on NarrativeQA (average length 16K; maximum length 32K) using Qwen3-4B as backbone.  
Each compression method's total latency can be divided into compression latency and inference latency. }
\label{tab:latency_results}
\begin{tabular}{l|ccccc}
\toprule
\textbf{Methods} & \multicolumn{5}{c}{\textbf{Compression Constraint}} \\
\cmidrule(lr){2-6}
& \textbf{2x} & \textbf{4x} & \textbf{8x} & \textbf{16x} & \textbf{32x} \\
\midrule
\multicolumn{6}{c}{\textbf{Compression Latency (s)}} \\
\midrule
EXIT & 303.61 & 300.86 & 301.30 & 302.45 & 302.06 \\
Provence & 2.12 & 2.12 & 2.12 & 2.13 & 2.11 \\
LongLLMLingua & 24.30 & 10.69 & 6.96 & 5.25 & 4.66 \\
\midrule
{\cellcolor[rgb]{0.925,0.957,1}}\textbf{RAM} & {\cellcolor[rgb]{0.925,0.957,1}}\textbf{0.10} & {\cellcolor[rgb]{0.925,0.957,1}}\textbf{0.09} & {\cellcolor[rgb]{0.925,0.957,1}}\textbf{0.09} & {\cellcolor[rgb]{0.925,0.957,1}}\textbf{0.09} & {\cellcolor[rgb]{0.925,0.957,1}}\textbf{0.08} \\
\midrule
\multicolumn{6}{c}{\textbf{Inference Latency (s)}} \\
\midrule
EXIT & 0.96 & 0.81 & 0.75 & 0.68 & 0.56 \\
Provence & 0.92 & 0.83 & 0.77 & 0.71 & 0.64 \\
LongLLMLingua & 0.87 & 0.68 & 0.65 & 0.59 & 0.48 \\
\midrule
{\cellcolor[rgb]{0.925,0.957,1}}\textbf{RAM} & {\cellcolor[rgb]{0.925,0.957,1}}\textbf{0.33} & {\cellcolor[rgb]{0.925,0.957,1}}\textbf{0.20} & {\cellcolor[rgb]{0.925,0.957,1}}\textbf{0.15} & {\cellcolor[rgb]{0.925,0.957,1}}\textbf{0.13} & {\cellcolor[rgb]{0.925,0.957,1}}\textbf{0.12} \\
\midrule
\multicolumn{6}{c}{\textbf{End-to-End Latency (s)}} \\
\midrule
EXIT & 304.57 & 301.67 & 302.05 & 303.13 & 302.62 \\
Provence & 3.04 & 2.95 & 2.89 & 2.84 & 2.75 \\
LongLLMLingua & 25.17 & 11.37 & 7.61 & 5.84 & 5.14 \\
\midrule
{\cellcolor[rgb]{0.925,0.957,1}}\textbf{RAM} & {\cellcolor[rgb]{0.925,0.957,1}}\textbf{0.43} & {\cellcolor[rgb]{0.925,0.957,1}}\textbf{0.29} & {\cellcolor[rgb]{0.925,0.957,1}}\textbf{0.24} & {\cellcolor[rgb]{0.925,0.957,1}}\textbf{0.22} & {\cellcolor[rgb]{0.925,0.957,1}}\textbf{0.20} \\
\midrule
Original Prompt & \multicolumn{5}{c}{1.23}\\
\bottomrule
\end{tabular}
\end{table}
\begin{equation}
\begin{split}
\text{FLOPs}_{\text{comp}} ={}& F_{\text{ParaEnc}}(Q, \mathbf{C}) \\
& + F_{\text{QueryAttn}}(Q, \mathbf{R}),
\end{split}
\end{equation}
where $F_{\text{ParaEnc}}(Q, \mathbf{C})$ denotes the cost of parallel encoding. Since each segment has length $L_{\text{seg}}$ and is processed independently, its complexity is $O(N \cdot L_{\text{seg}}^2) = O(L_{\text{seg}} \cdot L_{\text{org}})$, which is substantially lower than the $O(L_{\text{org}}^2)$ cost of full-sequence encoding (note $N \cdot L_{\text{seg}} = L_{\text{org}}$ and $L_{\text{seg}} \ll L_{\text{org}}$). Meanwhile, $F_{\text{QueryAttn}}(Q, \mathbf{R})$ (i.e., the cost of query-guided attention over $N$ relevance scores) incurs only $O(N)$ computational overhead, which is linear in the number of segments.

\paragraph{Decoding Stage.} Assuming the generated answer has length $L_a$, decoding requires $L_a$ forward passes. The FLOPs of the $i$-th pass depend on the compressed context length $L_c$ and query length $L_q$:
\begin{equation}
\text{FLOPs}_{i}^{\text{forward}} = F_{\text{Decoder}}(\mathbf{M}, Q, i),
\end{equation}
where $\mathbf{M}$ denotes the mixed input (retained tokens and skimmed vectors). The total FLOPs are therefore:
\begin{equation}
\text{FLOPs} = \sum_{i=1}^{L_a} \text{FLOPs}_{i}^{\text{forward}} + \text{FLOPs}_{\text{comp}}.
\end{equation}
Empirical results show that under the same 32$\times$ compression constraint, RAM achieves significantly lower end-to-end latency compared to task-relevant baselines (see Table~\ref{tab:latency_results}).

\begin{figure}[htb]
    \centering
    \includegraphics[width=1\linewidth]{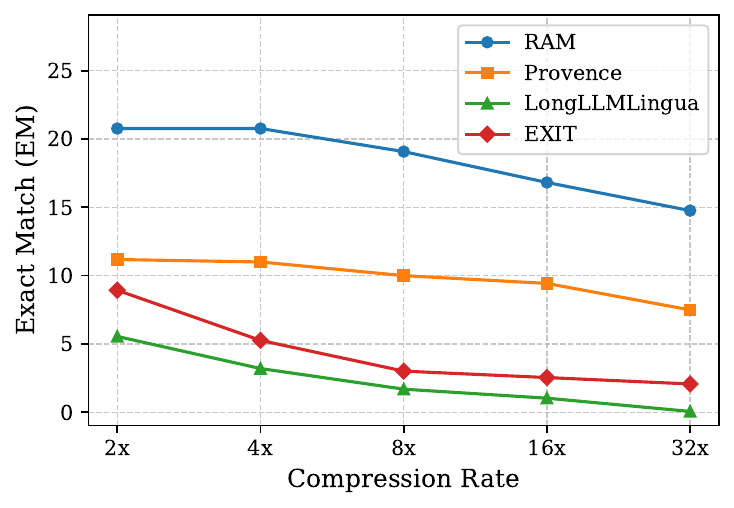}
    \caption{Performance under different compression rates on NarrativeQA. All methods use Qwen3-4B as the backbone.}
    \label{fig:diff_comp_rate}
\end{figure}

\subsection{Robustness Across Compression Rates (RQ3)}
\label{subsec:diff_comp_rate}
As shown in Figure~\ref{fig:diff_comp_rate}, RAM consistently maintains a clear advantage across different compression rates (from 2x to 32x), indicating that RAM can effectively adapt to varying compression levels via a single run and exhibits strong robustness. In contrast, baseline methods such as Provence, LongLLMLingua, and EXIT exhibit a steady decline in Exact Match (EM) scores as compression becomes more aggressive, suggesting their sensitivity to high compression ratios. Notably, while all methods use Qwen3-4B as the backbone, RAM’s performance remains relatively stable, demonstrating its resilience and practical utility for real-world applications requiring dynamic compression.

\begin{figure*}[htb]
    \centering
    \includegraphics[width=0.8\linewidth]{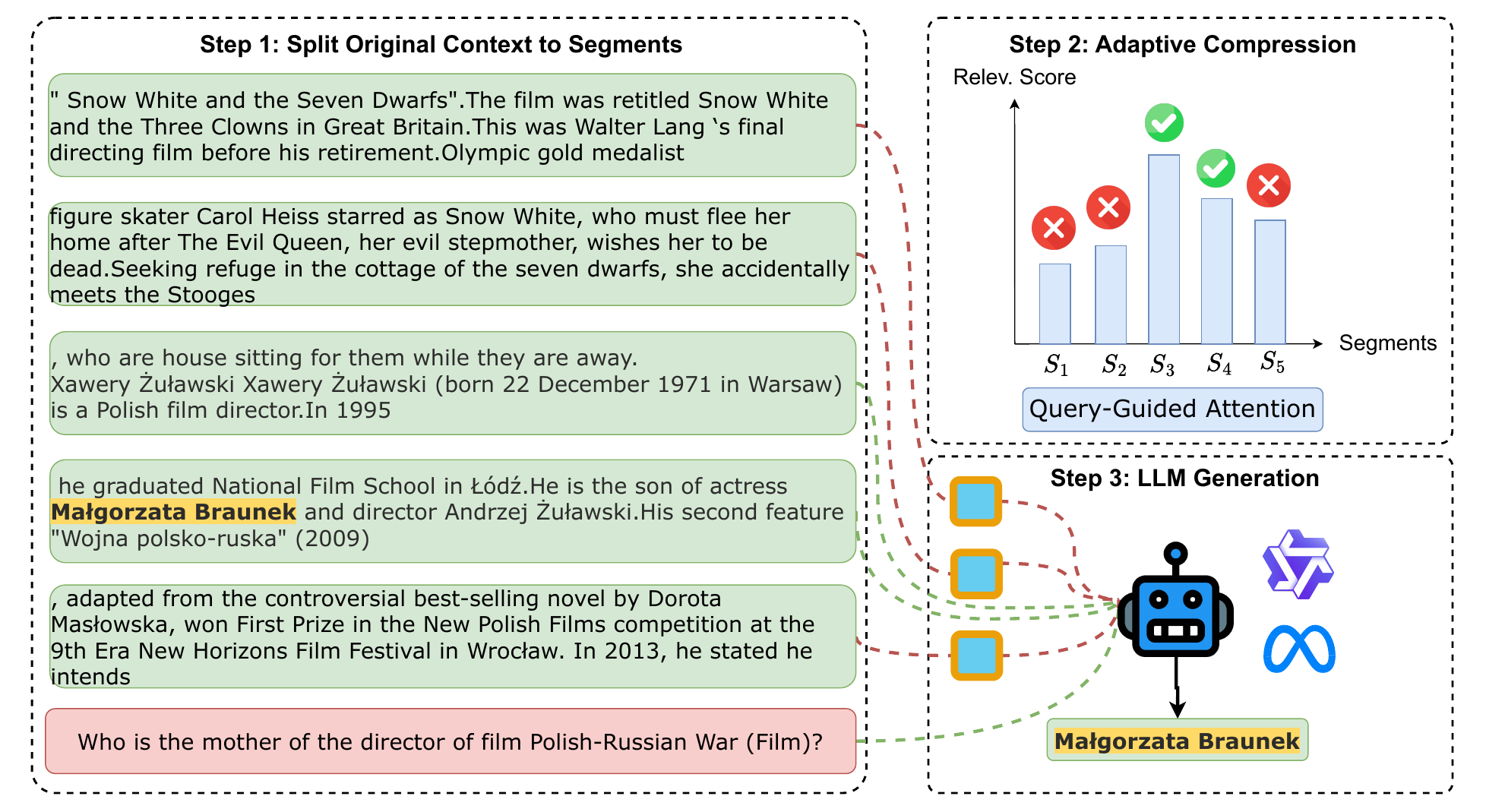}
    \caption{A case study from 2WikiMQA dataset.}
    \label{fig:case_study}
\end{figure*}

\begin{table}[htb]
\centering
\small
\caption{Ablation study on NaturalQuestions, 2WikiMQA under 8x compression constraint using Qwen3-4B as backbone.}
\begin{tabular}{ l | c c c c}
\toprule
\multirow{2}{*}{\textbf{Methods}} & 
  \multicolumn{2}{c}{\textbf{NaturalQA}} & 
  \multicolumn{2}{c}{\textbf{2WikiMQA}} \\
\cmidrule(lr){2-3} \cmidrule(lr){4-5}
& \textbf{EM} & \textbf{F1} & \textbf{EM} & \textbf{F1} \\
\midrule
{\cellcolor[rgb]{0.925,0.957,1}}\textbf{Default} & {\cellcolor[rgb]{0.925,0.957,1}}\textbf{62.41} & {\cellcolor[rgb]{0.925,0.957,1}}\textbf{57.14} & {\cellcolor[rgb]{0.925,0.957,1}}\textbf{39.63} & {\cellcolor[rgb]{0.925,0.957,1}}\textbf{45.24} \\
\midrule
\emph{w/o} Skimming & 58.38 & 53.95 & 37.19 & 42.10 \\
\emph{w/o} Close Reading & 46.40 & 45.97 & 34.28 & 39.31 \\
\emph{w/} AP Skimming & 59.54 & 54.10 & 37.21 & 42.30 \\
\emph{w/o} Contrast Learning & 49.27 & 46.03 & 33.29 & 37.42 \\
\bottomrule
\end{tabular}
\label{tab:ablation}
\end{table}
\subsection{Ablation Study (RQ4)}
To address RQ4, we conduct three ablation studies using Qwen3-4B as the backbone model to examine the contribution of each component in RAM to its overall performance:  
(1) The \textit{w/o} Skimming variant discards segments marked for compression after parallel encoding and adaptive selection, without generating any compressed tokens. (2) The \textit{w/o} Close Reading variant uniformly applies skimming to all segments. (3) The \textit{w/} AP Skimming variant replaces the query-guided skimming compression for each segment with simple Average Pooling (AP).  
(4) The \textit{w/o} Contrastive Learning variant removes the contrastive learning term from the total training objective.
Removing any component leads to a clear drop of the metric, demonstrating the necessity and effectiveness of each component. Discarding compressed tokens inevitably results in greater loss of key information, thereby degrading performance. Replacing query-aware compression with average pooling diminishes RAM’s sensitivity to query-relevant content within compressed segments, diluting key information in the compressed representation. Removing the contrastive loss weakens the model’s ability to distinguish between segments that require close reading or skimming, potentially leading to over-compression of important content and consequent performance degradation.

\subsection{Case Study}
RAM computes the relevance of each segment to the query using query-guided attention and applies close reading to highly relevant segments and skimming to less relevant ones. As shown in Figure~\ref{fig:case_study}, for the query ``Who is the mother of the director of film \textit{Polish-Russian War}?'' (the answer is \textit{Małgorzata Braunek}), segments $S_3$ and $S_4$ are processed with \emph{close reading} due to their high relevance, while the remaining segments undergo \emph{skimming}.

\section{Conclusion}
Inspired by human reading strategies, we propose RAM, a task-aware parallel context compression framework. RAM adaptively compresses query-irrelevant context segments (skimming) while preserving important parts (close reading). We further introduce a contrastive objective over the query and multiple context segments to better distinguish regions needing close reading from those suitable for skimming, improving compression quality. Extensive experiments demonstrate that RAM consistently outperforms strong baselines, delivering better effectiveness with substantial efficiency gains.

\section*{Limitations}
Although RAM demonstrates strong performance across various long-context benchmarks and compression rates, along with high efficiency and good interpretability, it has certain limitations. Specifically, the data labels used for contrastive learning on NarrativeQA are generated by the Qwen3-235B-A22B-Instruct. While this model provides high-precision labels, they are not guaranteed to be fully accurate. Nevertheless, as shown in Table~\ref{tab:ablation}, contrastive learning based on these labels remains effective.

\bibliography{main}

\appendix

\clearpage
\section{Impact of Segment Size}
\label{apx:impact_of_seg}

As shown in Table~\ref{tab:segment_size_ablation}, we conduct experiments to investigate the impact of varying segment sizes on RAM's performance. Under a 4x compression constraint with Qwen3-4B-Instruct as the backbone, we observe that the model consistently achieves strong and stable results across all three QA benchmarks. The low standard deviations demonstrate minimal performance variance across different segment granularities, highlighting the robustness and reliability of RAM in practical deployment scenarios.

\begin{table}[htb]
\centering
\fontsize{7.8}{9}\selectfont
\caption{Experimental results with varying segment sizes on three QA benchmarks using Qwen3-4B-Instruct as backbone under 4x compression constraint.}
\label{tab:segment_size_ablation}
\begin{tabular}{c | c c c c c c}
\toprule
\multirow{2}{*}{\textbf{Seg. Size}} & 
  \multicolumn{2}{c}{\textbf{2WikiMQA}} & 
  \multicolumn{2}{c}{\textbf{HotpotQA}} & 
  \multicolumn{2}{c}{\textbf{NarrativeQA}} \\
\cmidrule(lr){2-3} \cmidrule(lr){4-5} \cmidrule(lr){6-7}
& \textbf{EM} & \textbf{F1} & \textbf{EM} & \textbf{F1} & \textbf{EM} & \textbf{F1} \\
\midrule
50   & 50.39 & 56.47 & 46.37 & 59.28 & 19.45 & 31.92 \\
100  & 48.84 & 54.15 & 47.27 & 59.64 & 18.98 & 31.94 \\
200  & 47.52 & 52.90 & 46.06 & 58.41 & 19.74 & 31.32 \\
\midrule
\textbf{Std.} & 1.44 & 1.81 & 0.63 & 0.63 & 0.38 & 0.35 \\
\bottomrule
\end{tabular}
\end{table}

\section{Prompt Template of Annotation}
\label{apx:prompt_temp}
We employ a prompt template (see Figure~\ref{fig:prompt_template}) to instruct Qwen3-235B-A22B-Instruct to label each segment in NarrativeQA as positive if it is helpful for generating the ground-truth answer, and negative otherwise.
\begin{figure*}[htb]
\begin{tcolorbox}[
    colback=mybgblue!20,
    colframe=myblue,
    coltitle=white,
    fonttitle=\bfseries,
    title=Prompt Template,
    overlay={
        \draw[myblue, dashed] 
        (frame.west) -- (frame.east) 
        node[pos=0.5, above, yshift=3mm, font=\small] {PP}
        node[pos=0.5, below, yshift=-3mm, font=\small] {RR};
    }
]
Given the following question and segment, determine whether the segment contains sufficient information to answer the question. Respond with only ``Yes'' or ``No''. \\
\\
Question: \{question\} \\ \\

Segment: \{context\} \\ \\

Answer:
\end{tcolorbox}
\caption{Prompt template of annotation used by Qwen3-235B-A22B-Instruct.}
\label{fig:prompt_template}
\end{figure*}



\section{Dataset Details}
\label{apx:dataset}

\paragraph{NaturalQuestions} 
This dataset consists of real queries issued to the Google search engine, paired with entire Wikipedia pages. It requires models to identify both a long answer (typically a paragraph) and a short answer (one or more entities or a boolean) within the document. The corpus contains 307,373 training examples, 7,830 development examples, and 7,842 test examples. It is widely utilized for evaluating open-domain question answering and machine reading comprehension under realistic information-seeking scenarios. \emph{We select the processed
version~\cite{DBLP:journals/tacl/LiuLHPBPL24,DBLP:conf/acl/CaoCLPHCS24} where each question has 20 related documents and only one of them contains the correct answer. The processed version has 2,655 test samples.}

\paragraph{HotpotQA} 
HotpotQA is a large-scale dataset designed for multi-hop reasoning over Wikipedia articles. Questions are constructed such that the answer can only be found by performing reasoning across multiple documents. It features 112,779 training samples and 7,405 validation samples. A distinguishing factor of this dataset is the requirement for models to provide supporting facts (sentences) to explain the reasoning process, which enhances the explainability of the question-answering systems. \emph{We use the validation set to evaluate model's performance.}

\paragraph{2WikiMQA} 
This dataset focuses on multi-hop question answering with structured evidence, specifically utilizing Wikipedia and Wikidata. It aims to minimize the presence of "reasoning shortcuts" by using templates to generate complex questions that require synthesizing information from up to four documents. The dataset includes 167,454 training instances, 12,576 validation instances, and 12,576 test instances. It includes four types of reasoning: compositional, inference, comparison, and bridge. \emph{We use the test set to evaluate model performance.}

\paragraph{NarrativeQA} 
NarrativeQA is designed to test deep understanding of entire stories rather than local surface-level matching. It contains questions based on complete books and movie scripts. Unlike datasets that rely on short snippets, this benchmark requires models to capture long-range dependencies. The collection includes 1,572 documents divided into 1,102 for training, 115 for validation, and 355 for testing. These documents correspond to 46,765 total question and answer pairs.\emph{We use the test set to evaluate the model's performance, filtering out test samples longer than 32K.}

\paragraph{MultiNews} 
As a large-scale multi-document summarization dataset, Multi-News consists of news articles and human-written summaries. Each sample contains a summary paired with multiple source articles from various news sites. The dataset provides 44,972 training examples, 5,622 validation examples, and 5,622 test examples. It is a standard benchmark for evaluating the ability of models to consolidate redundant information and resolve conflicting details across diverse sources. \emph{We use the test set to evaluate model performance.}

\section{Ethic Statement}
This paper introduces RAM, an efficient and interpretable context compression framework inspired by human reading behavior. It combines parallelizable close reading and query-guided skimming to retain key information while reducing redundancy. The data and models used in this work are sourced from publicly available benchmarks and open-source platforms under appropriate licenses. While our method may influence how long-context LLMs are deployed, it does not introduce new ethical risks beyond those already present in existing context compression approaches. Thus, no additional ethical concerns require specific attention.

\section{Language Model Usage Statement}
In drafting this paper, we use a large language model to assist with academic writing. Specifically, we use it to improve wording, organization, and overall readability, including edits to the description of our methods and to the exposition of mathematical derivations. The scientific contributions of this work, including its key ideas, experimental setup, and reported results, are conceived, executed, and verified by the authors.



\end{document}